# A NONPARAMETRIC BAYESIAN APPROACH FOR SPOKEN TERM DETECTION BY EXAMPLE QUERY

*Amir Hossein Harati Nejad Torbati and Joseph Picone*

College of Engineering, Temple University
Philadelphia, Pennsylvania, USA
`Amir.harati@gmail.com, joseph.picone@temple.edu`

## Abstract

State of the art speech recognition systems use data-intensive context-dependent phonemes as acoustic units. However, these approaches do not translate well to low resourced languages where large amounts of training data is not available. For such languages, automatic discovery of acoustic units is critical. In this paper, we demonstrate the application of nonparametric Bayesian models to acoustic unit discovery. We show that the discovered units are correlated with phonemes and therefore are linguistically meaningful.

We also present a spoken term detection (STD) by example query algorithm based on these automatically learned units. We show that our proposed system produces a P@N of 61.2% and an EER of 13.95% on the TIMIT dataset. The improvement in the EER is 5% while P@N is only slightly lower than the best reported system in the literature.

## 1. Introduction

Acoustic unit discovery is a critical issue in many speech recognition tasks where there are limited resources available. Though traditional context-dependent phone models perform well when there is ample data, automatic discovery of acoustic units (ADU) offers the potential to provide good performance for resource deficient languages with complex linguistic structures (e.g., African click languages). Since only a small fraction of the world's 6,000 languages are currently supported by robust speech technology, this remains a critical problem.

Most approaches to automatic discovery of acoustic units [1]-[3] do this in two steps: segmentation and clustering. Segmentation is accomplished using a heuristic method that detects changes in energy and/or spectrum. Similar segments are then clustered using an agglomerative method such as a decision tree. Advantages of this approach include the potential for higher performance than that obtained using traditional linguistic units, and the ability to automatically discover pronunciation lexicons.

In this paper, we propose the use of a nonparametric Bayesian (NPB) model for automatically discovering acoustic units in speech recognition. In our formulation of the problem, the number of acoustic units is unknown. One brute force approach to this problem is to exhaustively search through a model space consisting of many possible parameterizations. However in an NPB model [4][5], the model complexity can be inferred directly from the data. Segmenting an utterance into acoustic units can be approached in a manner similar to that used in speaker diarization, where the goal is to segment audio into regions that correspond to a specific speaker. Fox et al. [6] used one state per speaker and demonstrated segmentation without knowing the number of speakers a priori. We have also previously reported on the use of a similar model for speech segmentation that achieves state of the art results [7]. This paper is continuation of our previous work and includes the application of two nonparametric hidden Markov models (HMMs) to the problem of acoustic unit discovery. These models include a Hierarchical Dirichlet Process HMM (HDPHMM) [6] and a Doubly Hierarchical Dirichlet Process HMM (DHDPHMM) [8][9].

Spoken term detection by example query (STD-EQ) is a problem that involves searching an audio corpus using a spoken query. One simple approach is to convert both the search query and the corpus into text form, often by using a conventional ASR system, and then to perform a text search. A slightly more flexible approach is to use a phonetic lattice to represent the search query and the corpus. However, converting the corpus and the query to text or to phonetic lattices involves using a state of the art speech recognizer that requires significant resources (e.g. dictionaries and transcribed training data). In this paper, we propose a new unsupervised algorithm based on ADUs that can approach state of the art performance amongst unsupervised approaches.

The rest of the paper is organized as follows: in Section 2, some background material related to nonparametric approaches used in the rest of the paper is introduced. In Section 33, the ADU transducer is presented. The STD-EQ problem is described in Section 4. In Section 5 some experimental results are discussed. These results include a comparison of ADU units with phonemes and a comparison of our STD by query algorithm to other unsupervised algorithms for the TIMIT dataset [10].

**Relationship to Previous Work:** In [7] we have used an NPB model for speech segmentation that achieves state of the art performance for unsupervised algorithms. The work presented here is a continuation of that work. Varadarajan et al. [11] proposed an algorithm to learn a speaker-dependent transducer that maps the acoustic observation to acoustic units. However, our proposed model is speaker independent and follows a different modeling approach.

Lee & Glass [12][13] proposed a nonparametric Bayesian model based on Dirichlet process mixtures (DPMs) that jointly segments the data and discovers the acoustic units. Our approach, however, discovers more homogenous units. We model each unit with a mixture of Gaussians while Lee & Glass use a 3-state HMM. Our ADU transducer also learns the sequential relationships between different units in the form of the probability of a transition from one to the next. Lee & Glass do not model these relationships.

## 2. Background

A Dirichlet process (DP) [14] is a discrete distribution that consists of a countably infinite number of probability masses. A DP is denoted by DP($\alpha$,$H$), and is defined as:

$$G = \sum_{k=1}^{\infty} \beta_k \delta_{\theta_k}, \quad \theta_k \sim H, \quad (1)$$

where $\alpha$ is the concentration parameter, $H$ is the base distribution, and $\delta_{\theta_k}$ is the unit impulse function at $\theta_k$, often referred to as an atom [15]. The weights $\beta_k$ are sampled through a stick-breaking construction [16] and are denoted by $\beta \sim GEM(\alpha)$. One of the applications of a DP is to define a nonparametric prior distribution on the components of a mixture model that can be used to define a mixture model with an infinite number of mixture components [15].

An HDP extends a DP to grouped data [17]. In this case there are several related groups and the goal is to model each group using a mixture model. These models can be linked using traditional parameter sharing approaches. One approach is to use a DP to define a mixture model for each group and to use a global DP, DP($\gamma$,$H$), as the common base distribution for all DPs [17]. An HDP is defined as:

$$\begin{aligned}
G_0 \mid \gamma, H &\sim DP(\gamma, H) \\
G_j \mid \alpha, G_0 &\sim DP(\alpha, G_0) \\
\theta_{ji} \mid G_j &\sim G_j \\
x_{ji} \mid \theta_{ji} &\sim F(\theta_{ji}) \quad \text{for } j \in J,
\end{aligned} \quad (2)$$

where $H$ provides a prior distribution for the factor $\theta_{ji}$, $\gamma$ governs the variability of $G_0$ around $H$ and $\alpha$ controls the variability of $G_j$ around $G_0$. $H$, $\gamma$ and $\alpha$ are hyperparameters of the HDP. We use a DP to define a mixture model for each group and use a global DP, DP($\gamma$,$H$), as the common base distribution for all DPs.

An HDPHMM [6] is an HMM with an unbounded number of states. The transition distribution from each state is modeled by an HDP. This lets each state have a different distribution for its transitions while the set of reachable states would be shared amongst all states. The definition for HDPHMM is given by [6]:

$$\begin{aligned}
\beta \mid \gamma &\sim GEM(\gamma) \\
\pi_j \mid \alpha, \beta &\sim DP(\alpha + \kappa, \frac{\alpha\beta + \kappa\delta_j}{\alpha + \kappa}) \\
\psi_j \mid \sigma &\sim GEM(\sigma) \\
\theta_{kj}^{**} \mid H, \lambda &\sim H(\lambda) \\
z_t \mid z_{t-1}, \{\pi_j\}_{j=1}^{\infty} &\sim \pi_{z_{t-1}} \\
s_t \mid \{\psi_j\}_{j=1}^{\infty}, z_t &\sim \psi_{z_t} \\
x_t \mid \{\theta_{kj}^{**}\}_{k,j=1}^{\infty}, z_t &\sim F(\theta_{z_t s_t}).
\end{aligned} \quad (3)$$

The state, mixture component and observation are represented by $z_t$, $s_t$ and $x_t$ respectively. The indices $j$ and $k$ are indices of the state and mixture components respectively. The base distribution that links all DPs together is represented by $\beta$ and can be interpreted as the expected value of state transition distributions. The transition distribution for state $j$ is a DP denoted by $\pi_j$ with a concentration parameter $\alpha$. Another DP, $\psi_j$, with a concentration parameter $\sigma$, is used to model an infinite mixture model for each state ($z_j$). The distribution $H$ is the prior for the parameters $\theta_{kj}$.

A DHDPHMM extends the definition of HDPHMM by allowing mixture components to be shared amongst different states [9]. The model definition for an ergodic DHDPHMM is given by:

$$\begin{aligned}
\beta \mid \gamma &\sim GEM(\gamma) \\
\pi_j \mid \alpha, \beta &\sim DP(\alpha + \kappa, \frac{\alpha\beta + \kappa\delta_j}{\alpha + \kappa}) \\
\xi \mid \tau &\sim GEM(\tau) \\
\psi_j \mid \sigma, \xi &\sim DP(\sigma, \xi) \\
\theta_{kj}^{**} \mid H, \lambda &\sim H(\lambda) \\
z_t \mid z_{t-1}, \{\pi_j\}_{j=1}^{\infty} &\sim \pi_{z_{t-1}} \\
s_t \mid \{\psi_j\}_{j=1}^{\infty}, z_t &\sim \psi_{z_t} \\
x_t \mid \{\theta_{kj}^{**}\}_{k,j=1}^{\infty}, z_t &\sim F(\theta_{z_t s_t}).
\end{aligned} \quad (4)$$

We have previously shown DHDPHMM can improve performance in problems such as acoustic modeling [9].

## 3. An ADU Transducer

The goal in speech segmentation is to map each acoustic observation into a segment and optionally label these segments. Our goal can be expressed as mapping a string of acoustic observations to a string of labels. In speech recognition, observations are vectors of real numbers (instead of symbols in text processing) and segment labels can be replaced with a vector that represents the posterior probability of a set of predefined symbols. This representation is called a posteriorgram [18].

A transducer specifies a binary relationship for a pair of strings [19]. Two strings are related if there is a path in the transducer that maps one string to the other. A weighted transducer also assigns a weight for each pair of strings [19]. Based on this definition our problem is to find a transducer that maps a string of acoustic features onto a string of units. It should be noted that based on this definition any HMM can be considered to be a transducer. We chose the term transducer here to emphasize the operation of converting acoustic observations into acoustic units. The problem can be further divided into two sub-problems: learning a transducer and decoding a string of observations into a string of units (or their equivalent posteriorgram representation).

Let's assume we already knew the acoustic units (e.g. phonemes) and have trained models for each unit (e.g. HMMs). One way to construct a transducer is to connect all these HMMs using an ergodic network. The final transducer can be some form of ergodic HMM. However, we don't have the units and the number of units in the data is unknown.

In [7] we used HDPHMM for speech segmentation. In [9] we introduced a DHDPHMM that allows sharing mixture components across states. These models can learn different structures including ergodic structures. Both of these models are good candidates to train a transducer. A C++ implementation of both algorithms that also includes DPM and HDP is available at [20].

We use an HDPHMM or DHDPHMM to train the transducer. Learning HDPHMM and DHDPHMM models is extensively discussed in [6][9]. Here we train the models in a completely unsupervised fashion. Unlike Lee & Glass [12][13] we don't utilize a speech/non-speech classifier and model everything including silence with one transducer. For read speech, this does not present any problems. However, for other domains such as conversational speech, it might be a problem, and in that case we can employ a speech/non-speech classifier as well. Training is executed by sequentially presenting utterances to the HDPHMM/DHDPHMM inference algorithm and iterating using Gibbs sampling.

For our transducer, state labels (or their posteriorgrams) are the output string. Since each state is modeled by a Gaussian mixture, the segments defined by this transducer are stationary and the discovered units are sub-phonetic. However, it should be noted that this limitation can be overcome by replacing each state (e.g. mixture model) with an HMM which transforms the model into a hierarchical HMM [21]. The resulting model can model dynamic segments.

Given a transducer and a string of observations the goal of the decoder is to find the most likely path through states of the transducer that implicitly maps the input string to the output string. This objective can be written as:

$$\arg\max_{s_1 s_2 \ldots s_M} P(s_1 s_2 \ldots s_M | o_1 o_2 \ldots o_N), \quad (5)$$

where $s_1, s_2, \ldots, s_M$ represent state labels and $o_1, o_2, \ldots, o_N$ represent observations. Alternately, we can also estimate the posteriorgram of the state sequence. To optimize (5) we can utilize the Viterbi algorithm [22]. The resulting transducer is the engine used to convert new acoustic observations into acoustic units.

## 4. Unsupervised STD by Example Query

Spoken term detection by query is a system that can recover data containing a word or phrase given a query example. An STD system can be built based on a complex state of the art speech recognizer and work either by acoustic or text queries. However, building such a system requires all the resources needed to build a state of the art speech recognizer including a lexicon, a language model and plenty of transcribed data. Moreover, if a given word does not exist in the lexicon we might never recover it using an ASR-based system.

An alternative approach is to use a phoneme sequence or some other low level equivalent. In this paper, we have used an ADU transducer for this goal. Our unsupervised STD-EQ algorithm is as follows:

1. Convert the target audio data using the ADU transducer into posteriorgrams.
2. For each query generate its posteriorgram representation using the transducer.
3. Use a subsequence dynamic time warping (DTW) algorithm [23] to obtain a score for each utterance.
4. Compare the final score for each utterance with a threshold and return it if the score is greater than the threshold.

DTW is used to align two sequences X and Y with different lengths. The distance between X and Y is defined as [23]:

$$DTW(X,Y) \quad C_{p^*}(X,Y) \\ = \min\{C_p(X,Y) | p \text{ is the warping path between } X \text{ and } Y\}, \quad (6)$$

where $C_p(X,Y)$ is defined as:

$$C_p(X,Y) = \sum_{i=1}^{L} c(x_{nl}, y_{ml}). \quad (7)$$

Note that in this formulation $c(x_{nl}, y_{ml})$ is an element of the cost matrix between $X$ and $Y$. Since $X$ and $Y$ are posteriorgrams, this cost is defined as the dot product between them [18]:

$$c(x,y) = -\log(x \cdot y). \quad (8)$$

The subsequence DTW algorithm computes the distance between two strings. The goal is to find a subsequence of $Y$ such as $Y(a^*, b^*)$ that minimizes the distance from $X$. Denoting the length of $Y$ as $M$, this objective can be expressed as:

$$(a^*, b^*) = \arg\min_{(a,b): 1 \le a \le b \le M} (DTW(X, Y(a,b))). \quad (9)$$

In our implementation we have replaced the strings with posteriorgrams and used the appropriate cost described in (8).

## 5. Experiments

In this section some experimental results are presented. First ADU units are compared with phonemes. Second we compare our STD-EQ algorithm with some other unsupervised algorithms on the TIMIT dataset [10]. TIMIT contains 6,300 utterances. We used the training subset to extract example queries and the test subset to search for the query. A standard 39-dimensional MFCC feature vector was used (12 MFCC plus energy and their first and second derivatives) to convert speech data into feature streams.

### 5.1. Relationship with Phonemes

It is important to explore the relationship between the ADUs and phonemes because we need to determine if the ADUs are linguistically meaningful. The first experiment involves aligning manually transcribed phonemes with ADUs. First, each utterance is passed through the transducer to generate the sequence of ADUs. Then these ADUs are aligned with manual transcriptions using timing information contained in the transcription. Finally, a confusion matrix is calculated.

A confusion matrix between 48 English phonemes and 251 ADU units is shown in Figure 1. A general correlation between ADUs and phonemes can be observed because the diagonal region of the matrix is heavily populated. However the mapping is not consistently one to one. Some of the ADUs align with multiple phonemes. These phonemes are generally similar phonemes. For example, we can see ADUs that are aligned with "sil" (silence) can also be aligned with "vcl" and "cl" models (both "vcl" and "cl" are special types of silence). ADUs aligned with "z" can also be aligned with "s". This is not surprising because "z" and "s" are similar acoustically and therefore confusable.

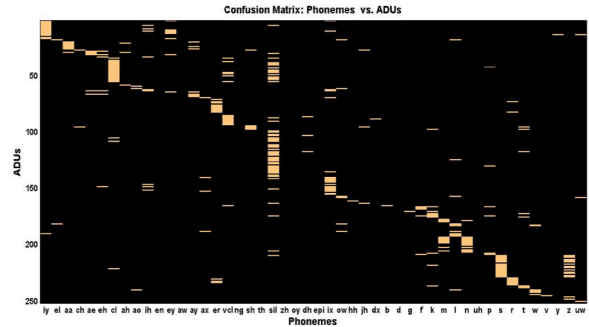

Figure 1: A confusion matrix that shows the relationship between ADUs and phonemes.

### 5.2. STD by Example Query

Table 2 shows the queries used to assess the quality of our ADU transducer. Since some words are similar (year vs. years), we do not count confusions between two words that share the same stem. We report the average precision of the top $N$ hits or $P@N$ [24] which is computed as the ratio of correct hits for top $N$ scores for a given keyword. The average $P@N$ is computed by averging the score for all keywords. We also reported the average equal error rate (EER). EER is the point on detection error tradeoff (DET) curve where the false acceptance error rate is equal to false rejection error rate. The reported EER is the average of EER for all keywords.

A comparison of the average $P@N$ and $EER$ is reported in Table 1 for TIMIT. The first row shows a system that utilizes a GMM to directly decode the posteriorgrams of the feature frames [18]. The second row shows the result of an algorithm based on a Deep Boltzmann Machine (DBM) [25]. The third row contains the results for the nonparametric Bayesian approaches described earlier [13]. Rows four and five contain the results of our ADU-based unsupervised systems.

We can see for both HDPHMM and DHDPHMM, the EER is lower than other unsupervised models (5% improvement relative to the best system) while the $P@N$ is only slightly lower than the NPM system in [13]. The primary reason that the HDPHMM transducer works better than the DHDPHMM transducer is the fact that for HDPHMM each state is modeled with a single Gaussian and this distribution is unique to that state, while for DHDPHMM all states share a pool of Gaussians. Each state can have more than one Gaussian associated with it, and this can make some states more confusable.

Row 6 shows the result of combining the output of HDPHMM and DHDPHMM-based models. The reason for this experiment was to investigate how much improvement can be obtained if we can use both systems. We have selected the best results of each system and therefore the result of row 6 is not reported as the result of our algorithm. However, it shows that HDPHMM and DHDPHMM are complementary. It also shows if we can combine the results of these two systems we can get very close to the results of a semi-supervised system shown in row 7 (e.g., using tied triphones in a conventional ASR system).

Table 2: A list of query terms used for the STD by example query task is shown.

| Query | No. Training | No. Test |
|---|---|---|
| age | 3 | 8 |
| warm | 10 | 5 |
| year | 11 | 5 |
| problem | 22 | 13 |
| artists | 7 | 6 |
| money | 19 | 9 |
| organizations | 7 | 6 |
| development | 9 | 8 |
| surface | 3 | 8 |
| children | 18 | 10 |

Table 1: A comparison of unsupervised approaches to STD by query is shown.

| System | P@N | EER |
|---|---|---|
| GMM [18] | 52.50% | 16.40% |
| DBM [24] | 51.10% | 14.70% |
| NPM [13] | 63.00% | 16.90% |
| DHDPHMM ADU | 56.21% | 14.33% |
| HDPHMM ADU | 61.20% | 13.95% |
| Combined HDPHMM/DHDPHHMM | 64.91% | 11.83% |
| Semi-supervised triphone [13] | 75.90% | 11.70% |

Table 3 shows some of the typical error pairs. It can be seen that for some cases we have a partial acoustic match between the search query and the retrieved word (e.g. message and age) and for others we have partial similarity (e.g. year and hear). These errors are clearly related to Figure 1 since they happen often for more confusable phonemes. From this table we can see our algorithm effectively finds all acoustically similar examples that might actually be pronounced in a manner similar to the target keyword in the dataset. Given the fact that ADU units are discovered automatically based on the acoustic similarities, this is an expected result.

### 6. Conclusions

In this paper we proposed the application of HDPHMM/DHDPHMM to the problem of learning acoustic units automatically. We have shown discovered ADU units have a meaningful relationship with phonemes. We have also proposed an unsupervised STD by example query algorithm based on these ADU units. We have shown our system can achieve state of the art results among unsupervised systems.

In the future, we intend to study an NPB model that models nonstationary units. As mentioned above, our current model assumes each unit can be represented with a Gaussian mixture. As a result our model discovers sub-phonetic units. If we can model each state of the HDPHMM with another HMM then this limitation would be eliminated.

Another direction for future research is to study the relationship between the quality of a search query and its textual form. It has been shown in [26] that in a standard STD system the quality of a search query can be predicated based on its spelling. However, it is an open question if this is also the case for the proposed algorithm.

### 7. Acknowledgements

This research was supported in part by the National Science Foundation through Major Research Instrumentation Grant No. CNS-09-58854.

Table 3: Error pairs for STD system.

| Query | Discovered |
|---|---|
| age | mess**age** |
| development | fulfil**lment** |
| year | d**ee**r |
| year | **here** |
| year | be**havior** |
| surface | **severe** |